\documentclass[runningheads]{llncs}
\usepackage[T1]{fontenc}
\usepackage{graphicx}
\usepackage{cite}
\usepackage{amsmath, amssymb}
\usepackage{amsfonts}
\usepackage{algorithmic}
\usepackage{graphicx}
\usepackage{textcomp}

\usepackage{booktabs}
\usepackage{multirow}
\usepackage{pifont}
\usepackage{bm}
\newcommand{\cmark}{\ding{51}}
\usepackage[table]{xcolor}
\usepackage[misc]{ifsym}

\begin{document}
\title{SalientFusion: Context-Aware Compositional Zero-Shot Food Recognition}

% \author{Anonymous submission 192}
\author{Jiajun Song\inst{1} \and
Xiaoou Liu\inst{1}\Letter }
%
% \authorrunning{Jiajun et al.}
\institute{$^1$Renmin University of China \\ \email{\{jiajun.song, xiaoou.liu\}@ruc.edu.cn} }

% \institute{}

\maketitle              % typeset the header of the contribution
\begin{abstract}

Food recognition has gained significant attention, but the rapid emergence of new dishes requires methods for recognizing unseen food categories, motivating Zero-Shot Food Learning (ZSFL). We propose the task of Compositional Zero-Shot Food Recognition (CZSFR), where cuisines and ingredients naturally align with attributes and objects in Compositional Zero-Shot learning (CZSL). However, CZSFR faces three challenges: (1) Redundant background information distracts models from learning meaningful food features, (2) Role confusion between staple and side dishes leads to misclassification, and (3) Semantic bias in a single attribute can lead to confusion of understanding. Therefore, we propose \underline{\textit{SalientFusion}}, a context-aware CZSFR method with two components: SalientFormer, which removes background redundancy and uses depth features to resolve role confusion; DebiasAT, which reduces the semantic bias by aligning prompts with visual features. Using our proposed benchmarks, CZSFood-90 and CZSFood-164, we show that \textit{SalientFusion} achieves state-of-the-art results on these benchmarks and the most popular general datasets for the general CZSL. The code is avaliable at \url{https://github.com/Jiajun-RUC/SalientFusion}.

\keywords{Compositional Zero-shot Learning  \and Food Recognition.}
\end{abstract}

\section{Introduction}

\label{introduction}

\begin{figure}[t]
 	\centering
 	\includegraphics[width=8cm]{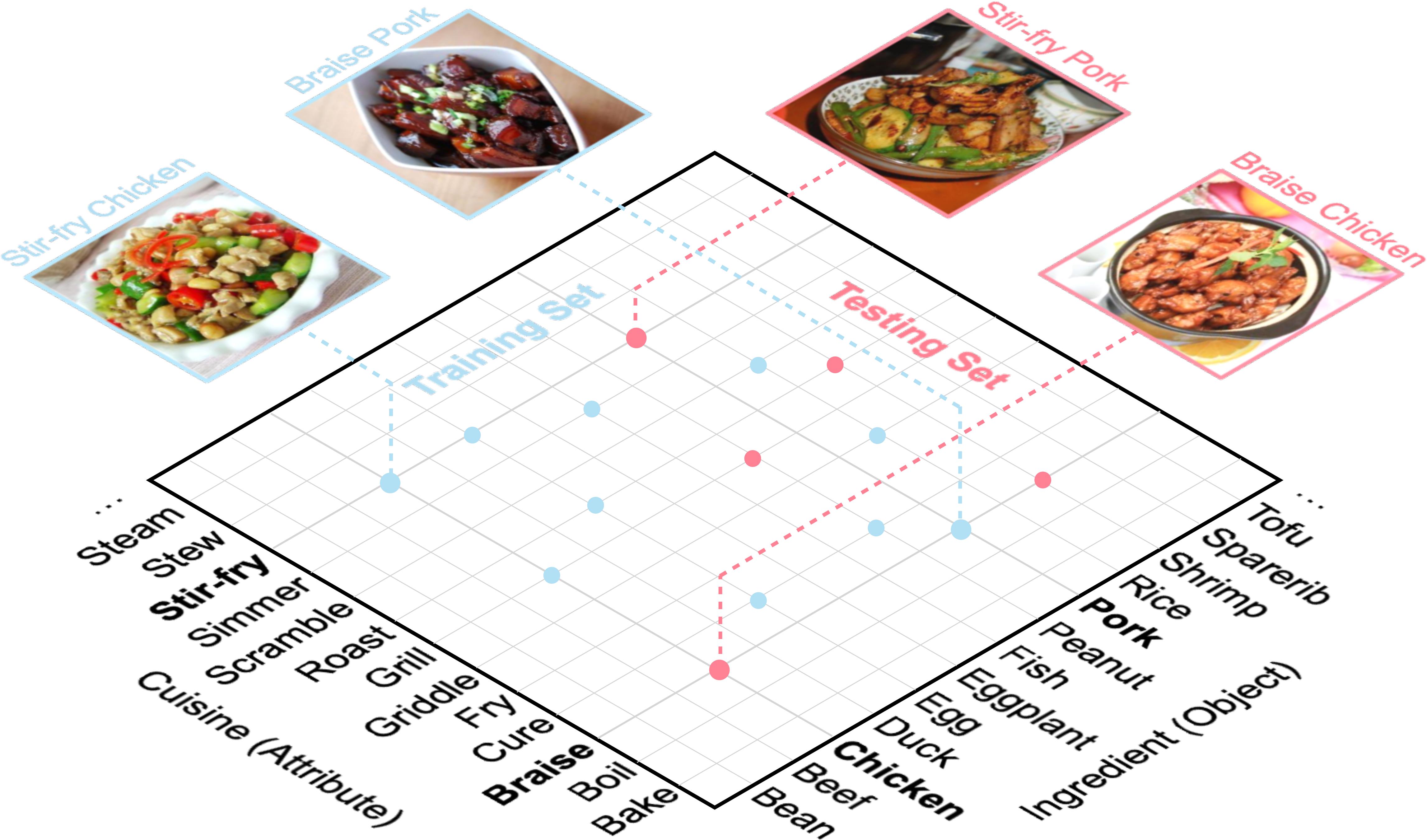}
 	\caption{An example diagram of compositional zero-shot tasks in the food domain: rows represent food cuisines, and columns represent food ingredients. Within the Cartesian product of cuisines and ingredients, the non-overlap between the training set and the testing set making it challenging to achieve out-of-distribution generalization.}
 	\label{Task}
    \vspace{-2em}
\end{figure}

Food is integral to daily life and underpins many societal processes. Despite significant advances in food recognition~\cite{min2023large,saranya2023comparative,zhou2021rwmf}—a cornerstone of food computing~\cite{min2019survey}—existing methods struggle with novel, unseen dishes. To overcome this, we introduce Zero-Shot Food Learning (ZSFL), a framework for recognizing food categories absent from training data. Given the limited research and lack of dedicated datasets for ZSFL~\cite{zhou2023seed,li2024ese}, we propose the task of Compositional Zero-Shot Food Recognition (CZSFR). In CZSFR, the inherent structure of food is exploited by aligning “cuisines” and “ingredients” with “attributes” and “objects” in  Compositional Zero-Shot Learning (CZSL). As illustrated in Fig.~\ref{Task}, our model is trained on seen compositions (e.g., “braise pork” and “stir-fry chicken”) and evaluated on unseen ones (e.g., “braise chicken” and “stir-fry pork”). This task not only directs future research in ZSFL but also highlights the critical role of compositional analysis in food recognition.

\begin{figure}[t]
 	\centering
 	\includegraphics[width=8cm]{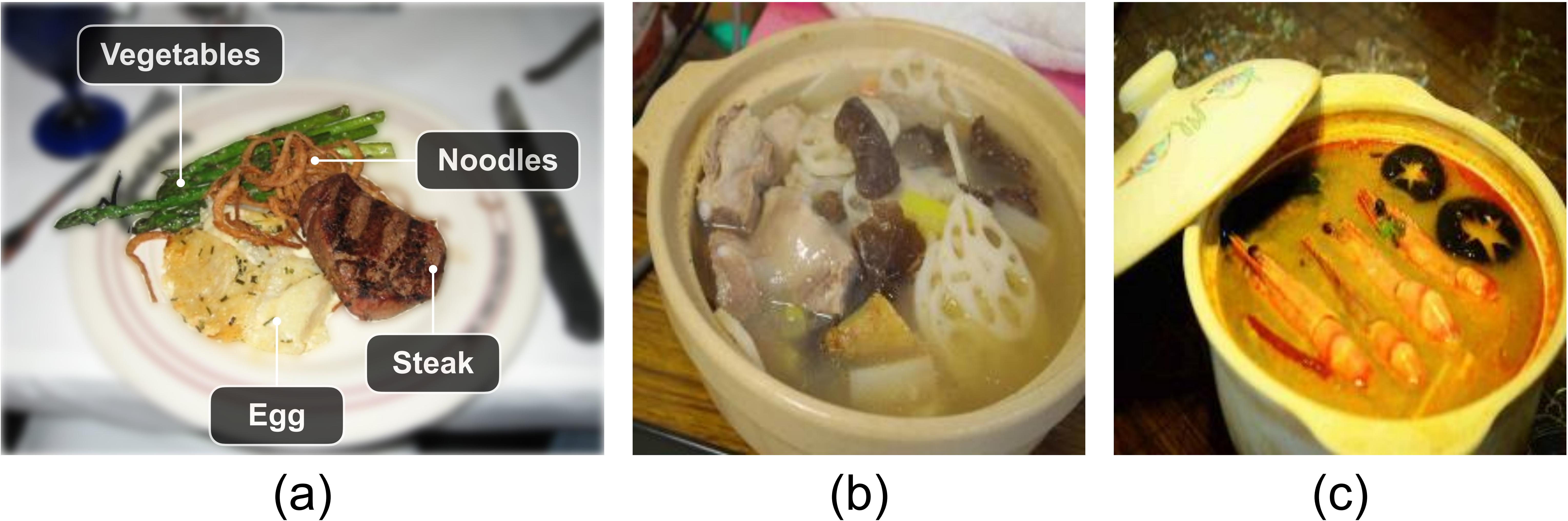}
 	\caption{Challenges of CZSFR. (a) shows the ``grilled beef'' image affected by redundant information and role confusion. (b) and (c) are ``stew sparerib'' and ``stew seafood'', representing different compositions within the same cuisine.}
 	\label{problem}
    \vspace{-2em}
\end{figure}

Existing CZSL methods face three key challenges in CZSFR: \textbf{(1) Redundant background information.} Food images often include irrelevant elements, such as plates and tables, which distract models from meaningful features. As shown in Fig.~\ref{problem} (a), Vision-Language Models (VLMs), for instance, may describe a grilled beef image as ``grilled beef on a round plate placed on a table'', where ``plate'' and ``table'' are irrelevant. Suppressing such distractions can help models focus on the food itself. \textbf{(2) Role confusion between the staple dish and the side dish.} Accurate classification requires distinguishing the staple dish from side dishes. As shown in Fig.~\ref{problem} (a), misidentifying a side dish, such as vegetables or noodles, as the staple dish can lead to errors. Emphasizing the staple’s role is crucial for precise categorization. \textbf{(3) Semantic bias in a single attribute can lead to confusion of understanding.} The single attribute like ``stew'' can differ in meaning across categories. As shown in Fig.~\ref{problem} (b) and (c), ``stew'' in stew sparerib aligns with ``braised'', while in stew seafood, it aligns with ``boil''. Addressing these biases requires aligning visual features with contextual prompts. By tackling these issues, models can better handle CZSFR and generalize to unseen combinations.

To address the above three identified challenges, we propose \textbf{\textit{SalientFusion}}, a context-aware method for CZSFR, comprising two key components: \textbf{SalientFormer} and \textbf{DebiasAT} module. First, SalientFormer tackles redundant background information and role confusion by combining image segmentation~\cite{zheng2024BiRefNet} and depth detection~\cite{Bochkovskii2024depth}. The segmentation removes irrelevant elements like plates and tables, while depth information can capture the volume and distance
of objects, which helps distinguish the staple dish from side dishes. These features are fused to produce a salient representation focused on meaningful food regions. Second, to address the semantic bias in a single attribute across different compositions, DebiasAT module aligns static text representations with salient visual features extracted by SalientFormer. By dynamically guiding text learning based on visual context, this module ensures more accurate semantic alignment and reduces attribute misinterpretation. Together, these components enable \textit{SalientFusion} to effectively address CZSFR’s unique challenges, redundant information, role confusion and semantic bias in a single attribute, setting the stage for improved generalization to unseen food compositions.

Finally, we re-divide and re-annotate two popular food image datasets, ETH Food-101~\cite{bossard2014food} and VireoFood-172~\cite{Chen-DIRCRR-MM2016}, to create two novel benchmarks, CZSFood-90 and CZSFood-164, tailored for the CZSFR task. Specifically, we replace the original category labels with compositions of cuisines and ingredients to align with the compositional nature of the task. To better evaluate model performance in practical scenarios, we introduce a real-world testing method. This approach comes from the observation that most novel food categories arise from discovering new ingredients or recombining existing ingredients in a new way. The real-world testing therefore evaluates models on unseen cuisine-ingredient compositions, simulating how new dishes emerge in real-world settings. Our proposed framework, \textit{SalientFusion}, achieves state-of-the-art (SOTA) performance on both benchmarks under closed-world and real-world testing settings. Additionally, \textit{SalientFusion} demonstrates its generalizability by achieving SOTA results on the MIT-States~\cite{isola2015discovering}, a benchmark widely used in the general CZSL task beyond food recognition. These results highlight the robustness and effectiveness of our method in addressing the challenges of CZSFR.

Overall, our main contributions can be summarized as follows:
\begin{itemize}
\item We are the first to explore CZSFR and extensively explore existing CZSL methods on food image datasets.

\item We introduce the \textit{SalientFusion} for CZSFR. The proposed SalientFormer and the DebiasAT module solve the problem of the redundant information, role confusion and semantic bias in a single attribute.

\item We benchmark the CZSFR with two food image datasets and propose the real-world testing method. The proposed \textit{SalientFusion} achieves the best performance on all food datasets and a widely-used general dataset.
\end{itemize}

\section{Related Work}

\label{related_work}

\begin{figure*}[t]
 	\centering
 	\includegraphics[width=12cm]{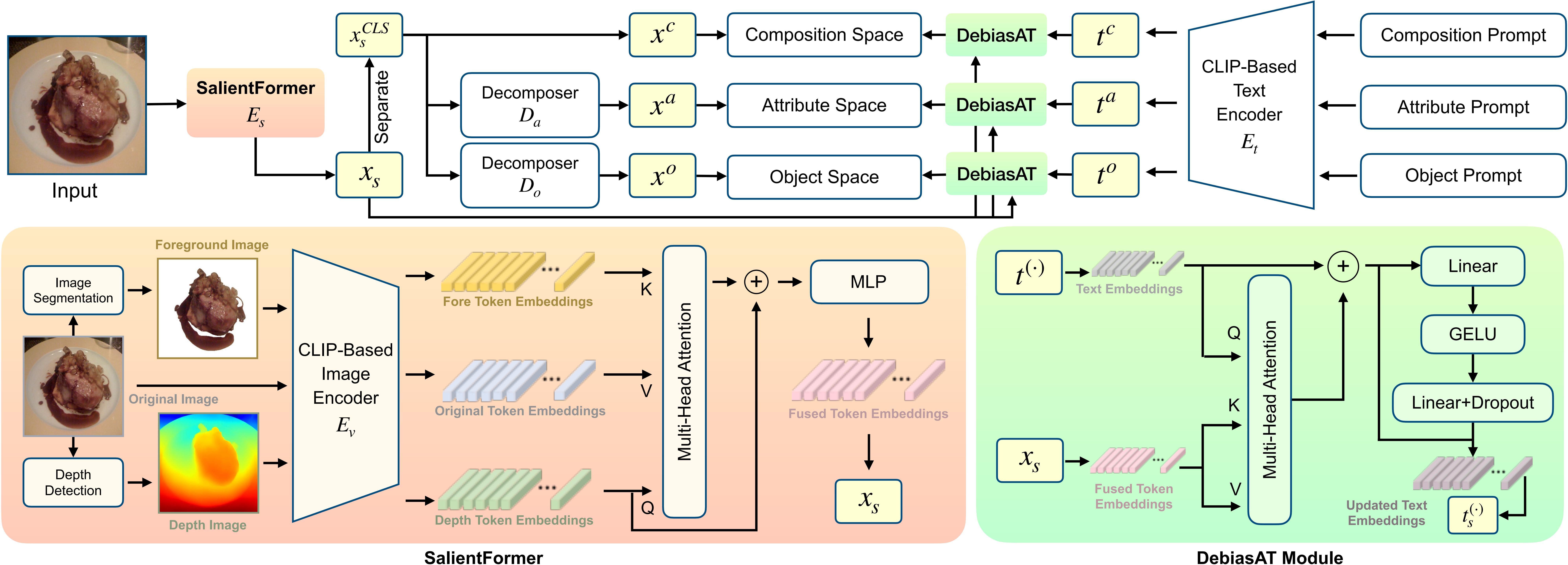}
 	\caption{\textbf{Architecture of \textit{SalientFusion}}. The SalientFormer extracts salient features via segmentation and depth detection and fuses them, while the DebiasAT module aligns text prompts with output fused features to address compositional challenges using CLIP-based encoders.}
 	\label{framework}
    \vspace{-2em}
\end{figure*}

\textbf{From ZSFL to Compositional Understanding.} Traditional Zero-Shot Food Learning (ZSFL) methods, whether generative (e.g., ESE-GAN~\cite{li2024ese}) or attribute-based (e.g., SeeDS~\cite{zhou2023seed}), typically treat food categories as monolithic entities. This holistic view limits their ability to generalize to novel dishes created by combining known cuisines and ingredients. Our work addresses this by shifting focus to Compositional Zero-Shot Food Recognition (CZSFR), which explicitly models the compositional nature of food.

\textbf{Compositional Zero-Shot Learning (CZSL).} Existing CZSL methods primarily fall into two camps. (1) Representation-based approaches like CompCos~\cite{li2022siamese} and SymNet~\cite{saini2022disentangling} aim to learn disentangled representations for primitives (e.g., attributes and objects). While effective at separating concepts, they often oversimplify the complex visual interaction between them. For instance, the appearance of ``fish'' is fundamentally altered when ``fried'' versus served as ``sashimi''. (2) VLM-based approaches have shown great promise by leveraging powerful pre-trained models like CLIP. Methods such as CSP~\cite{nayak2023softprompts} and DFSP~\cite{lu2023decomposed} learn to generate compositional text prompts to guide recognition. However, by processing the image holistically, these methods are susceptible to background clutter and ``role confusion'' — failing to correctly associate an attribute with its corresponding object in a multi-object scene (e.g., associating ``fried'' with ``chicken'' and not the side of ``rice'').

\textbf{Discussion.} The limitations of prior work highlight a critical gap: a lack of explicit context-awareness in the composition process. \textit{SalientFusion} directly addresses this and moves beyond holistic image processing by first identifying the salient region of the primary ingredient. Then, it fuses the features of this specific region with contextual cues derived from the cuisine primitive. This mechanism is designed to mitigate the effects of background noise and prevent role confusion, thereby enabling a more robust recognition of unseen food compositions.

% CZSL~\cite{hou2021fabricated, kato2018compositional, yang2022causal} employs two primary strategies to infer unseen compositions. The first assumes that seen and unseen compositions share the same attribute and object scopes, predicting primitive labels (attributes and objects) separately and then combining them into composition labels. For instance, Liu et al.~\cite{liu2023primitives} leverage contextual semantics to independently predict attributes and objects for composition inference, while Li et al.~\cite{li2023distilled} employ reversed attention to disentangle attributes and objects. Second, many works~\cite{li2022siamese,nagarajan2018attributes,saini2022disentangling} directly aligns images with textual labels in a shared space to predict compositions by identifying the most similar pairs. For example, Nagarajan et al.~\cite{nagarajan2018attributes} construct a composition space by simulating visual changes caused by attribute-object interactions, and Anwaar et al.~\cite{anwaar2022variational} enhance composition learning with a composition graph. 
\section{Method}

\label{method}

\subsection{Task Formulation}
Given the state set \( \mathcal{A} = \{a_1, a_2, \dots, a_m\} \) and object set \( \mathcal{O} = \{o_1, o_2, \dots, o_n\} \) as primitive concepts, where $m$ and $n$ represent the cardinality of \(\mathcal{A}\) and \(\mathcal{O}\). The compositional label space \( \mathcal{C} \) is defined as their Cartesian product: \( \mathcal{C} = \mathcal{A} \times \mathcal{O} \), and $p$ is the cardinality of \(\mathcal{C}\). The seen and unseen composition sets, denoted as \( \mathcal{C}_{\text{se}} \) and \( \mathcal{C}_{\text{us}} \), are two disjoint subsets of \( \mathcal{C} \), satisfying \( \mathcal{C}_{\text{se}} \cap \mathcal{C}_{\text{us}} = \emptyset \). To train a model that assigns compositional labels from a target set \( \mathcal{C}_{\text{tgt}} \) to input images, a training set \( \mathcal{T} = \{(\bm{x_i}, c_i) \mid \bm{x} \in \mathcal{X}, c \in \mathcal{C}_{\text{se}}\} \) is provided, where \( \mathcal{X} \) denotes the image space. The target set is defined as \( \mathcal{C}_{\text{tgt}} = \mathcal{C}_{\text{se}} \cup \mathcal{C}_{\text{us}} \), considering only the known composition space.

\subsection{SalientFusion}
% \vspace{-2mm}
In this paper, we propose a context-aware method, \textit{SalientFusion}, including SalientFormer for extracting image features and DebiasAT for correcting the semantic bias in text embeddings. \textit{SalientFusion} is illustrated in Fig.~\ref{framework}.

\textbf{Learning Visual Representations.} Rather than simply using a frozen image encoder as in previous methods~\cite{lu2023decomposed,zhang2024learning}, we argue that extracting more salient and distinctive features from food images is crucial for improving the performance. First, the input food image $\bm{x} \in \mathbb{R}^{\text{H} \times \text{W} \times \text{C}}$ is fed into the proposed image encoder SalientFormer $E_s:  \bm{x}\mapsto \bm{x}_s$ to generate the image feature  $\bm{x}_s \in \mathbb{R}^{(\text{N} + 1) \times d}$, where $d$ represents the dimensionality of the cross-modal latent space and N is the number of patches. Moreover, the global image descriptor $\bm{x}^{\text{CLS}}_s \in \mathbb{R}^{d}$ is separated from $\bm{x}_s$. By keeping  $\bm{x}^{\text{CLS}}_s$ as the composition visual representation $\bm{x}^{c}$, we introduce the attribute decomposer $D_{a}: \mathbb{R}^{d} \to \mathbb{R}^{d}$ and object decomposer $D_{o}:\mathbb{R}^{d} \to \mathbb{R}^{d}$ to independently derive the attribute and object visual features, $\bm{x}^{a}$ and $\bm{x}^{o}$, which are denoted as $\bm{x}^{i} = D_{i}(\bm{x}^c),\quad i \in \{a,o\}$. The implementation of $D_a$ and $D_o$ relies on two separate Multi-Layer Perceptrons (MLPs), which extract distinct visual features for attribute and object to enable the cross-modal alignment across different branches.

\textbf{Learning Text Representations.} In the recent VLM-based method~\cite{lu2023decomposed}, composition prompts are typically restricted to retrieving information related to seen compositions. To overcome this limitation, \textit{SalientFusion} introduces distinct prompts for the attribute, object, and composition branches, which account for the independence of each primitive by designing dedicated prompts for attributes and objects. Building on the approach of previous works~\cite{huang2024troika}, \textit{SalientFusion} incorporates independent, learnable prompt prefixes for each branch. For any given attribute-object pair $c_{i,j} = \langle a_i, o_j \rangle$, the prompts are defined as:
\begin{equation}
\begin{aligned}
\bm{P}^{a}_i &= [\bm{p}^{a}_1, \dots, \bm{p}^{a}_r, \bm{v}^{a}_i],\\
\bm{P}^{o}_j &= [\bm{p}^{o}_1, \dots, \bm{p}^{o}_r, \bm{v}^{o}_j],\\
\bm{P}^{c}_{i,j} &= [\bm{p}^{c}_1, \dots, \bm{p}^{c}_r, \bm{v}^{c}_i, \bm{v}^{c}_j],
\end{aligned}
\end{equation}
where ${\bm{p}_1, \dots, \bm{p}_r}$ represent learnable prefix tokens, $r$ is the length of prefix tokens and $\bm{v}$ represents vocabulary tokens~\cite{nayak2023softprompts}. Then these prompts are fed into the text encoder $E_t:\mathbb{R}^{r \times d} \to \mathbb{R}^{d}$ to obtain text representations, formulated as $\bm{t}_i^a = E_t\left(\bm{P}_i^a\right)$, $\bm{t}_j^o = E_t\left(\bm{P}_j^o\right)$, and $\bm{t}_{i,j}^c = E_t\left(\bm{P}_{i,j}^o\right)$.

\textbf{Training.} With the visual representations and text representations defined for each branch, the probabilities of assigning the labels for attribute $a_i$, object $o_j$, and composition $c_{i,j}$ to the given image can be independently calculated as:
\begin{equation}
\begin{aligned}
p(a_i \mid \bm{x}) &= \frac{\exp(\bm{x}^a \cdot \bm{t}_i^a/\tau)}{\sum_{k=1}^{m} \exp(\bm{x}^a \cdot \bm{t}_k^a/\tau)},\\
p(o_i \mid \bm{x}) &= \frac{\exp(\bm{x}^{o} \cdot \bm{t}_i^{o}/\tau)}{\sum_{k=1}^{n} \exp(\bm{x}^{o} \cdot \bm{t}_k^{o}/\tau)},\\
p(c_{i,j} \mid \bm{x}) &= \frac{\exp(\bm{x}^{c} \cdot \bm{t}_{i,j}^{c}/\tau)}{\sum_{k=1}^{p} \exp(\bm{x}^{c} \cdot \bm{t}_k^{c}/\tau)},\\
\end{aligned}
\end{equation}
where $\tau \in \mathbb{R}$ is the pre-trained temperature parameter from CLIP. In each branch, the cross-entropy loss encourages the model to explicitly recognize the corresponding semantic role, described as:
\begin{equation}
\begin{aligned}
\mathcal{L}^k = -\frac{1}{|\mathcal{X}|} \sum_{x \in \mathcal{X}} \log p(k|\bm{x}), \quad k \in \{\text{a}, \text{o}, \text{c}\}.
\end{aligned}
\end{equation}
The final loss $\mathcal{L}$ is denoted as:
\begin{equation}
\mathcal{L} = \beta^a\mathcal{L}^a + \beta^o\mathcal{L}^o + \beta^c\mathcal{L}^c,
\end{equation}
where $\beta^a$, $\beta^o$ and $\beta^c \in \mathbb{R}$ are weights to balance the influence of different losses.

\subsection{SalientFormer}

SalientFormer is proposed to extract more salient image features to solve the problem of the redundant information and the role confusion. Unlike existing methods~\cite{huang2024troika}, which directly input the original image into a CLIP-based image encoder, we take a different approach by first exploring simple yet effective image segmentation~\cite{zheng2024BiRefNet} and depth detection techniques~\cite{Bochkovskii2024depth} that are easy to implement to preprocess the input. Specifically, the input image \( \bm{x} \) is processed through the image segmentation module \( T_s \) and depth detection module \( T_d \), generating the foreground image \( \bm{x}_f \) and depth mapping image \( x_d \), respectively:
\begin{equation}
T_s : \bm{x} \mapsto \bm{x}_f, \quad T_d : \bm{x} \mapsto \bm{x}_d.
\end{equation}
Both outputs retain the original dimensions of \( \bm{x} \), ensuring compatibility with downstream tasks. Then each image of the tuple $(\bm{x},\bm{x}_f,\bm{x}_d)$ is first disassembled into N fixed-sized patches and it is then projected into N tokens which serve as the input to the ViT and $\rm N = HW / P^2$, where $\rm (P, P)$ is the resolution of each patch. The position information is incorporated by adding a learnable $\rm 1-D$ positional vector to input tokens. An extra CLS token is added to the beginning of the input sequence to aggregate the information from all tokens to generate the global representation of the image. The CLIP-based image encoder $E_v: \mathbb{R}^{\text{H} \times \text{W} \times \text{C}} \to \mathbb{R}^{(\text{N} + 1) \times d^{\text{in}}}$ processes the token sequence using blocks based on self-attention, where $d^{\text{in}}$ represents the dimensionality of each visual token. Then the token sequence is passed through a linear layer, defined by the parameters $\mathbf{W}_g \in \mathbb{R}^{d^{\text{in}} \times d}$. Therefore, output token embeddings $\bm{\hat{x}}_o, \bm{\hat{x}}_f,\bm{\hat{x}}_d$ are denoted as
\begin{equation}
\bm{\hat{x}}_i = E_v(\bm{x}_{i})\mathbf{W}_g^\top, \quad i \in \{\text{o}, \text{f}, \text{d}\},
\end{equation}
where $\bm{\hat{x}}_o, \bm{\hat{x}}_f,\bm{\hat{x}}_d \in \mathbb{R}^{(\text{N}+1) \times d}$. Then, queries $\mathbf{Q}$, keys $\mathbf{K}$, and values $\mathbf{V}$ are defined as:
\begin{equation}
\begin{aligned}
    \mathbf{Q} &\leftarrow \alpha \bm{\hat{x}}_d + (1-\alpha) \bm{\hat{x}}_f, \\
    \mathbf{K} &\leftarrow \alpha \bm{\hat{x}}_f + (1-\alpha) \bm{\hat{x}}_d, \\
    \mathbf{V} &\leftarrow \bm{\hat{x}}_o,
\end{aligned}
\end{equation}
where $\mathbf{Q},\mathbf{K},\mathbf{V} \in \mathbb{R}^{(\text{N}+1) \times d}$, 
$\alpha \in [0,1]$ is a gating parameter that balances the contributions of the two token embeddings. Moreover, we explore the Multi-Head Attention mechanism (MHA) $M: \mathbb{R}^{(\text{N} + 1) \times d} \to \mathbb{R}^{(\text{N} + 1) \times d} $ to fuse these token embeddings, which are denoted as 
\begin{equation}
\bm{x}_s = M(\mathbf{Q},\mathbf{K},\mathbf{V} )
\end{equation}
where $\bm{x}_s$ is output fused token embeddings.

\begin{table}[t]
\centering
\caption{Dataset Statistics}
\fontsize{7pt}{8pt}\selectfont
\setlength{\tabcolsep}{0.6mm}{
\begin{tabular}{@{}lcccccccccccc@{}}
\toprule
\multicolumn{1}{c}{\multirow{3}{*}{\textbf{Dataset}}} & \multirow{3}{*}{$\mathcal{A}$} & \multirow{3}{*}{$\mathcal{O}$} & \multicolumn{2}{c}{\multirow{2}{*}{\textbf{Training}}} & \multicolumn{2}{c}{\multirow{2}{*}{\textbf{Validation}}} & \multicolumn{4}{c}{\textbf{Testing}}\\
% \cmidrule(lr){8-11} 
&&&&&&& \multicolumn{2}{c}{Closed}&\multicolumn{2}{c}{Real} \\ 
\cmidrule(lr){4-5} \cmidrule(lr){6-7} \cmidrule(lr){8-9} \cmidrule(lr){10-11}
& & & $\mathcal{C}$ & $\mathcal{X}$ & $\mathcal{C}$ & $\mathcal{X}$ & $\mathcal{C}$ & $\mathcal{X}$ & $\mathcal{C}$ & $\mathcal{X}$\\ 
\midrule
CZSFood-90&11&62&32&28.8k&16&9.6k&16&9.6k&58&51.6k \\
CZSFood-164&16&118&54&32.1k&20&8.1k&20&8.6k &110&64.3k  \\
MIT-States~\cite{isola2015discovering}& 115 & 245 & 1262 & 30.3k & 600 & 10.4k & 800 & 13.0k& - & - \\
\bottomrule
\end{tabular}}
\label{data}
\vspace{-2em}
\end{table}

\subsection{DebiasAT Module}

Despite the cross-modal understanding capabilities of VLMs, the semantic bias remains prevalent, especially when visual content varies across different primitives~\cite{huang2024troika}. To mitigate this issue, the proposed DebiasAT module leverages fused token embeddings extracted by the SalientFormer to refine static text representations. Specifically, given an input text representation $\bm{t} \in \mathbb{R}^d$ and the fused token embeddings \(\bm{x}_s \in \mathbb{R}^{(\text{N}+1) \times d} \) from the SalientFormer, the Multi-Head Attention (MHA) mechanism is employed to integrate the two modalities and compute the alignment, defined as:
\begin{equation}
\begin{aligned}
    \bm{\tilde{t}} &= \mathcal{N}\big(\bm{t} + M(\bm{t}, \bm{x}_s, \bm{x}_s)\big), \quad \bm{\tilde{t}} \in \mathbb{R}^d,\\
    \bm{t}^\prime &= \mathcal{N}\big(\bm{\tilde{t}} + \mathcal{F}_{\text{FFN}}(\bm{\tilde{t}})\big), \qquad \ \ \ \bm{t}' \in \mathbb{R}^d,
\end{aligned}
\end{equation}
where $M$ represents the Multi-Head Attention mechanism, and $\mathcal{F}_{\text{FFN}}: \mathbb{R}^d \to \mathbb{R}^d$ denotes the Feed-Forward Network. $\mathcal{N}: \mathbb{R}^d \to \mathbb{R}^d$ represents the layer normalization, which is applied to stabilize training and enhance convergence.

The updated text representation $\bm{t}_s\in \mathbb{R}^d$ is computed as:
\begin{equation}
    \bm{t}_s = \bm{t} + \lambda \cdot \bm{t}',
\end{equation}
where \(\lambda \in [0,1]\) is a gate controlling the extent of adjustment. This formulation ensures that the refined text representation \(\bm{t}_s\) aligns more closely with the salient visual content while maintaining consistency with the original representation $\bm{t}$.

\begin{table}[t]

	\centering
    
	\caption{Benchmarks comparison with SOTA methods. 
Highest and second best are highlighted in \textbf{bold} and \underline{underline}, respectively.}

    \fontsize{7pt}{8pt}\selectfont
	\setlength{\tabcolsep}{1.2mm}{

	\begin{tabular}{lcccccccc}

	     &\multicolumn{4}{c}{\textbf{CZSFood-90}} &\multicolumn{4}{c}{\textbf{CZSFood-164}} \\ 

         \bottomrule

		   Methods & S & U & HM & AUC & S & U & HM & AUC \\ \hline

            \multicolumn{9}{c}{\color{gray}\textit{Closed-World Testing}}\\ \hline

		\multicolumn{1}{l}{TMN \cite{purushwalkam2019task} }       &70.1&20.2&23.8&10.9&76.9&25.3&29.7&15.4\\

		\multicolumn{1}{l}{SymNet \cite{li2020symmetry}} &70.1&26.3&27.2&14.0&75.5&30.3&32.9&18.0\\

	    \multicolumn{1}{l}{SCEN \cite{li2022siamese}}   &72.9&22.7&26.1&12.9&73.7&23.2&29.1&14.7 \\

	    \multicolumn{1}{l}{CANet \cite{wang2023learning}}  &74.2&26.6&30.1&15.4&78.9&37.1&39.4&24.3 \\

	    \multicolumn{1}{l}{CLIP \cite{radford2021learning} } &26.8&59.3&29.6&11.4&19.6&46.8&20.4&7.2 \\

		\multicolumn{1}{l}{COOP \cite{zhou2022learning}} &63.4&\underline{64.8}&53.0&38.2&74.0&63.3&55.7&41.2\\

        \multicolumn{1}{l}{ESE-GAN \cite{li2024ese}} &75.6&29.1&32.6&17.8&74.3&25.7&30.2&22.1 \\

	    \multicolumn{1}{l}{Troika \cite{huang2024troika}} &\underline{97.5}&62.2&\underline{66.2}&\underline{56.6}&\textbf{96.7}&\underline{70.3}&\underline{70.6}&\underline{63.7} \\

            \rowcolor{gray!15}

	    \multicolumn{1}{l}{\textbf{SalientFusion}} &\textbf{97.8}{$\pm$\tiny .2}&\textbf{69.5}{$\pm$\tiny .8}&\textbf{68.0}{$\pm$\tiny .5}&\textbf{61.7}{$\pm$\tiny .5}&\underline{96.4}{$\pm$\tiny .3}&\textbf{78.3}{$\pm$\tiny .6}&\textbf{74.4}{$\pm$\tiny .3}&\textbf{70.9}{$\pm$\tiny .7} \\ \hline

        \multicolumn{9}{c}{\color{gray}\textit{Real-World Testing}}\\ \hline

        \multicolumn{1}{l}{TMN \cite{purushwalkam2019task} }       &70.0&5.7&8.8&3.0&75.3&3.4&5.5&1.9\\

		\multicolumn{1}{l}{SymNet \cite{li2020symmetry}} &70.1&4.2&6.7&2.2&75.3&3.5&5.9&2.1\\

	    \multicolumn{1}{l}{SCEN \cite{li2022siamese}}   &71.6&3.2&15.3&1.3&72.2&4.0&6.6&2.2 \\

	    \multicolumn{1}{l}{CANet \cite{wang2023learning}}  &74.8&4.5&7.5&2.7&79.0&4.9&8.1&3.1 \\

	    \multicolumn{1}{l}{CLIP \cite{radford2021learning} } &26.4&21.4&18.4&5.4&18.3&16.7&14.1&2.7 \\

		\multicolumn{1}{l}{COOP \cite{zhou2022learning}} &91.3&\underline{23.6}&30.6&18.8&76.8&\underline{24.0}&31.0&17.1\\

        \multicolumn{1}{l}{ESE-GAN \cite{li2024ese}} &74.9&13.2&15.8&12.5&73.2&14.1&15.7&11.3 \\

	    \multicolumn{1}{l}{Troika \cite{huang2024troika}} &\underline{96.4}&22.9&\underline{34.3}&\underline{21.0}&\underline{97.0}&23.3&\underline{34.0}&\underline{21.3} \\

            \rowcolor{gray!15}

	    \multicolumn{1}{l}{\textbf{SalientFusion}} &\textbf{96.9}{$\pm$\tiny .3}&\textbf{27.5}{$\pm$\tiny .8}&\textbf{39.0}{$\pm$\tiny .6}&\textbf{25.4}{$\pm$\tiny .5}&\textbf{97.6}{$\pm$\tiny .2}&\textbf{26.5}{$\pm$\tiny .3}&\textbf{38.0}{$\pm$\tiny .5}&\textbf{24.8}{$\pm$\tiny .2} \\ \hline

	\end{tabular}}

	\label{sota}
\vspace{-2em}
\end{table}

\section{Experiment}

\label{experiment}

\subsection{Datasets}

We create two new benchmarks, CZSFood-90 and CZSFood-164, by re-annotating the ETH Food-101~\cite{bossard2014food} and VireoFood-172~\cite{Chen-DIRCRR-MM2016} datasets. Our goal is to redefine each food category as a (cuisine, ingredient) composition. The re-annotation process is systematic. To define the cuisine primitives, we leverage a large language model to generate an exhaustive list of common cooking methods. For the ingredient primitives, we manually identify the one-to-two most salient ingredients for each dish. Each resulting (cuisine, ingredient) composition, such as (grill, chicken), is then meticulously verified against authoritative culinary reference guides to ensure accuracy. This process yields 90 compositions for CZSFood-90 and 164 for CZSFood-164 (see Tab.\ref{data}). As per standard practice\cite{nayak2023softprompts,wang2023learning}, we partition the compositions into training, validation, and closed-world test sets. A separate real-world test set, comprising unseen compositions (e.g., novel ingredient combinations), is created to specifically evaluate generalization. To further demonstrate robustness, we also evaluate our method on the MIT-States~\cite{isola2015discovering} and the UT-Zappos~\cite{yu2014fine} benchmark.

\subsection{Metrics and Implementation Details}

\textbf{Metrics.} We adopt the evaluation protocol from prior studies~\cite{huang2024troika}, which involves a calibration bias to balance the prediction scores between seen and unseen pairs during testing. To evaluate overall performance across both seen and unseen pairs, we calculate the area under the curve (AUC) and identify the point with the highest harmonic mean (HM) of seen and unseen accuracies. Additionally, we report the best seen accuracy (S) and the best unseen accuracy (U).

\textbf{Implementation Details.} Following~\cite{huang2024troika}, we utilize the pre-trained CLIP ViT-L/14 model as the image encoder and a transformer as the text encoder. We train and evaluate models on an NVIDIA 3090 GPU. We adopt the \texttt{Adam} optimizer with an initial learning rate of $5*10^{-5}$ and a weight decay of $10^{-5}$.

\begin{table}[t]

	\centering

	\caption{Ablation study of SalientFormer. $o,f,d$ represent the origin, foreground and depth image, respectively. }
    \fontsize{7pt}{8pt}\selectfont
	\setlength{\tabcolsep}{1.7mm}{

	\begin{tabular}{ccccccc|cccc}

	      \multicolumn{3}{c}{\textbf{Branch}}&\multicolumn{4}{c}{\textbf{CZSFood-90}} &\multicolumn{4}{c}{\textbf{CZSFood-164}}\\ 
\bottomrule
		  \textit{o} &\textit{f}&\textit{d}   & S & U & HM & AUC & S & U & HM & AUC  \\ \hline
          
            \multicolumn{11}{c}{\color{gray}\textit{Closed-World Testing}}\\ \hline

	    \cmark&& &97.5&62.2&66.2&56.6&\textbf{96.7}&70.3&70.6&63.7 \\
        
            \cmark&&\cmark &\underline{97.7}&\underline{67.4}&\underline{67.9}&\underline{60.4}&96.1&75.4&71.4&67.0 \\ 
            \cmark&\cmark& &97.6&63.4&63.3&55.8&\underline{96.4}&\underline{76.6}&\underline{74.1}&\underline{69.7} \\

            \rowcolor{gray!15} \cmark&\cmark&\cmark &\textbf{97.8}&\textbf{69.5}&\textbf{68.0}&\textbf{61.7}&\underline{96.4}&\textbf{78.3}&\textbf{74.4}&\textbf{70.9} \\ \hline

            \multicolumn{11}{c}{\color{gray}\textit{Real-World Testing}}\\ \hline

            \cmark&& &\underline{96.4}&22.9&34.3&21.0&97.0&23.3&34.0&21.3 \\
        
            \cmark&&\cmark &96.3&\underline{24.5}&\underline{35.1}&\underline{22.3}&\underline{97.3}&\underline{24.2}&\underline{34.1}&\underline{22.0} \\ 
            \cmark&\cmark& &\underline{96.4}&24.0&{35.1}&22.0&\textbf{97.6}&22.3&32.0&20.3 \\

            \rowcolor{gray!15}\cmark&\cmark&\cmark &\textbf{96.9}&\textbf{27.5}&\textbf{39.0}&\textbf{25.4}&\textbf{97.6}&\textbf{26.5}&\textbf{38.0}&\textbf{24.8} \\ \hline

	\end{tabular}}

	\label{ablation_SFFN}
% \vspace{-1em}
\end{table}

\begin{table}[t]

	\centering

	\caption{Ablation study on DebiasAT (\%). }
    \fontsize{7pt}{8pt}\selectfont
	\setlength{\tabcolsep}{1.6mm}{

	\begin{tabular}{ccccc|cccc}

	      &\multicolumn{4}{c}{\textbf{CZSFood-90}} &\multicolumn{4}{c}{\textbf{CZSFood-164}}\\ 
\bottomrule
		    \textit{SalientFusion} & S & U & HM & AUC & S & U & HM & AUC  \\ \hline
          
            \multicolumn{9}{c}{\textit{\color{gray}Closed-World Testing}}\\ \hline
          
	    w/o DebiasAT &97.0&67.2&67.5&60.9&95.8&72.1&72.2&65.6 \\       
            \rowcolor{gray!10} w/ DebiasAT  &\textbf{97.8}&\textbf{69.5}&\textbf{68.0}&\textbf{61.7}&\textbf{96.4}&\textbf{78.3}&\textbf{74.4}&\textbf{70.9} \\ \hline

            \multicolumn{9}{c}{\textit{\color{gray}Real-World Testing}}\\ \hline

            w/o DebiasAT &96.2&26.7&38.9&24.7&\textbf{97.6}&26.2&37.9&24.3 \\
            \rowcolor{gray!10} w/ DebiasAT &\textbf{96.9}&\textbf{27.5}&\textbf{39.0}&\textbf{25.4}&\textbf{97.6}&\textbf{26.5}&\textbf{38.0}&\textbf{24.8} \\ \hline

	\end{tabular}}

	\label{ablation_SCN}
\vspace{-1em}
\end{table}

\subsection{Results Analysis}

\textbf{Performance Comparison on Food Datasets.} In Tab.~\ref{sota}, we report the results of both closed-world and real-world testing. \textit{SalientFusion} consistently outperforms state-of-the-art methods across two food datasets. Compared with the baseline \textit{Troika}, \textit{SalientFusion} achieves improvements in HM by +1.8\%, +3.8\%, +4.7\%, and +0.7\%, and in AUC by +5.1\%, +7.2\%, +4.4\%, and +0.9\% across the two datasets and testing conditions. \textit{SalientFusion} achieves the best unseen accuracy, highlighting its strong generalization ability. Moreover, we observe that CLIP-based methods significantly outperform non-CLIP methods in CZSFR tasks, further demonstrating the robustness of VLMs in food task. These results emphasize the importance of leveraging pre-trained visual-text alignment for compositional generalization.

\textbf{Ablation Study on SalientFormer and DebiasAT.} In Tab.~\ref{ablation_SFFN}, we evaluate the impact of removing one or more specific branches in SalientFormer, highlighting the importance of each branch. The results show that the origin image alone performs the worst, while adding the depth image branch improves performance by helping the model focus on salient regions. Furthermore, the best performance is achieved when all three branches are combined, as the foreground image further enhances the model’s focus on salient areas. This improvement is consistent across both datasets, demonstrating the robustness of the results. Tab.~\ref{ablation_SCN} validates the effectiveness of the proposed DebiasAT module, further boosting performance by addressing semantic bias in text prompts.

\begin{figure}[t]
 	\centering
 	\includegraphics[width=8.3cm]{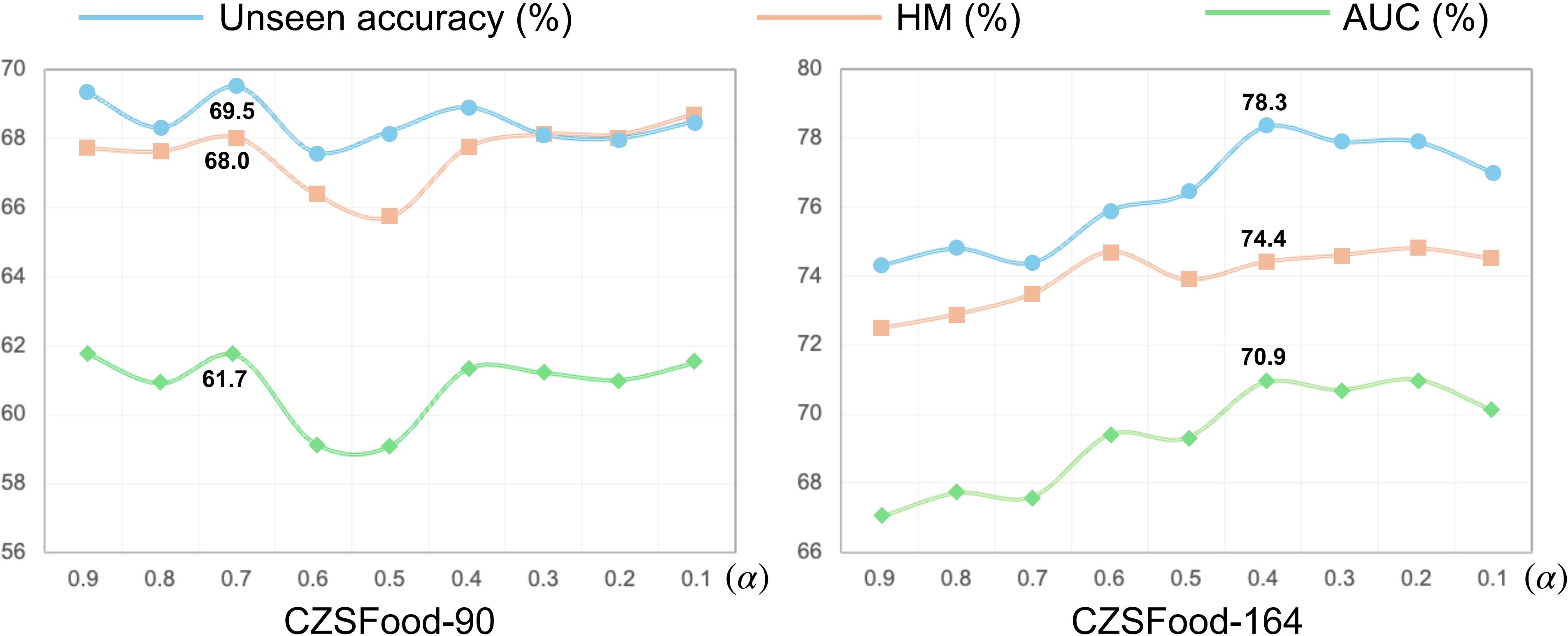}
 	\caption{The influence of the gate $\alpha$ on different datasets.}
 	\label{hyper}
        % \vspace{-5mm}
\end{figure}

\begin{figure}[t]
 	\centering
 	\includegraphics[width=8.8cm]{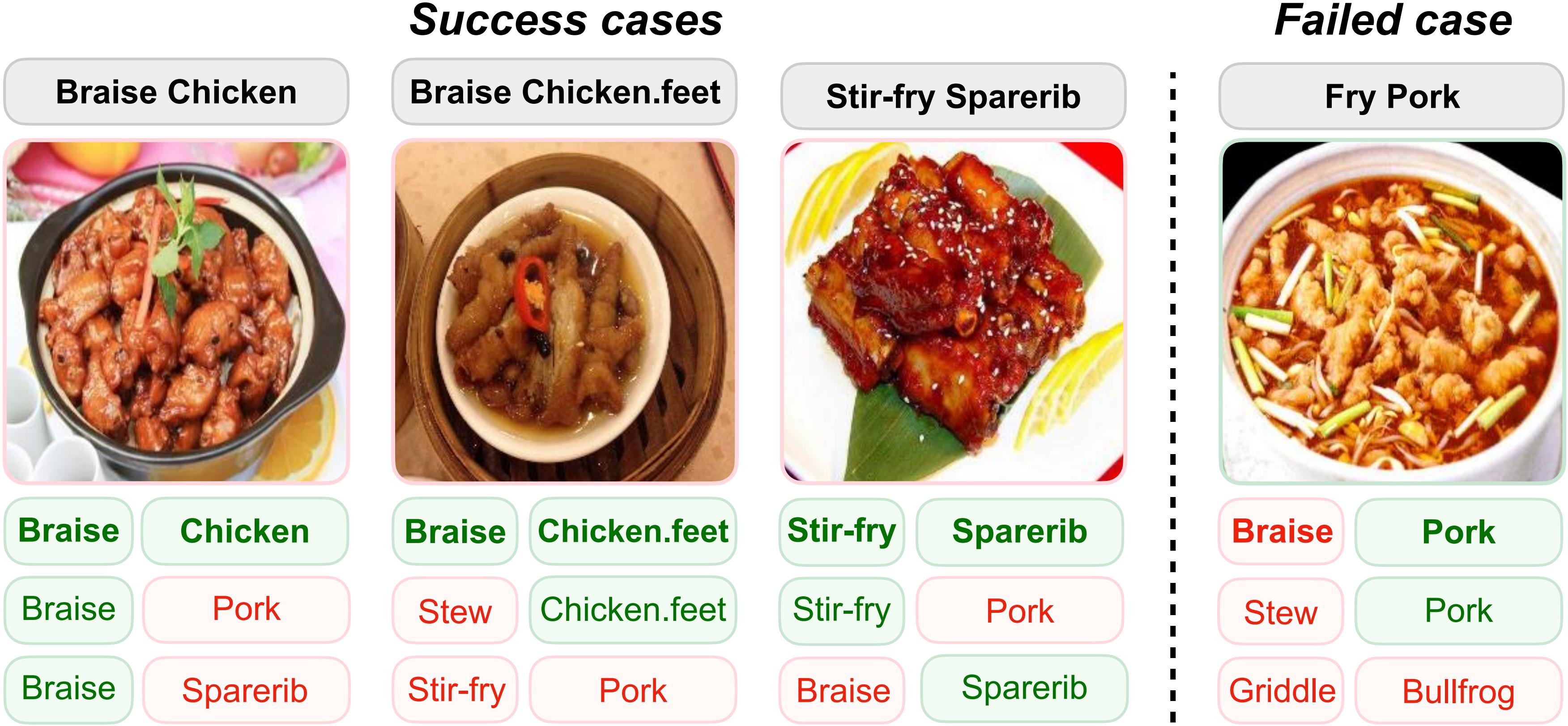}
 	\caption{Top-3 predictions of \textit{SalientFusion} for some instances. Black, green and red represents the ground truth, the correct prediction and the wrong one.}
 	\label{visual}
    \vspace{-1em}
\end{figure}
\textbf{Impact of Hyper-parameters.} We investigate the effect of different hyper-parameter settings, \textit{i.e.}, gate $\alpha$ in SalientFormer. The influence of $\alpha$ is shown in Fig.~\ref{hyper}, we observe that $\alpha = 0.7$ results in the best performance on CZSFood-90 and $\alpha = 0.4$ on CZSFood-164. The larger $\alpha$ on CZSFood-90 is due to the nature of Western cuisine, where dishes usually have well-defined staple and side dishes. In such cases, depth becomes critical in optimization, serving as the key query feature to resolve role confusion. Conversely, CZSFood-164, which represents Chinese cuisine, predominantly features secondary processing of mixed ingredients, minimizing the impact of role confusion, resulting in a smaller $\alpha$.

\textbf{Qualitative Analysis.} We present qualitative results in Fig.~\ref{visual} to analyze our model's behavior. The success cases, such as ``Braise Chicken'' and ``Stir-fry Sparerib'', demonstrate that \textit{SalientFusion} effectively recognizes correct cuisine-ingredient compositions by learning to associate salient visual patterns with their corresponding primitives. For the failure case, our model's top predictions are (Braise, Pork) and (Stew, Pork), indicating that it correctly identifies the ingredient (``Pork'') but confuses the cuisine. This error arises from the high visual ambiguity between certain cooking methods. The dish's saucy, browned appearance shares strong visual cues with ``braised'' or ``stewed'' preparations, while the distinctive features of ``frying'' are less prominent. This analysis reveals a key limitation: while our salient-region focus leads to robust object recognition, the model can be confounded by the inherent visual similarities among fine-grained cuisine primitives. This suggests that purely visual approaches may have a performance ceiling on such nuanced tasks. Future work could address this by incorporating multi-modal context, such as recipe text, to disambiguate visually similar cooking styles.

\begin{table}[t]

	\centering
    
    % \color{red}
	\caption{Comparison with SOTA on MIT-States and UT-Zappos (\%). }
    
    \fontsize{7pt}{8pt}\selectfont
	\setlength{\tabcolsep}{1.2mm}{

	\begin{tabular}{lcccccccc}

    &\multicolumn{4}{c}{\textbf{MIT-States}} &\multicolumn{4}{c}{\textbf{UT-Zappos}}
    \\ \bottomrule
		     Methods & S & U& HM & AUC & S & U& HM & AUC  \\ \bottomrule

		\multicolumn{1}{l}{TMN \cite{purushwalkam2019task} }   &20.2&20.1&13.0&2.9&58.7&60.0&45.0&29.3\\

		\multicolumn{1}{l}{SymNet \cite{li2020symmetry}} &24.4&25.2&16.1&3.0&49.8&57.4&40.4&23.4\\

	    \multicolumn{1}{l}{SCEN \cite{li2022siamese}}   &29.9&25.2&18.4&5.3&63.5&63.1&47.8&32.0 \\

	    \multicolumn{1}{l}{CANet \cite{wang2023learning}}  &29.0&26.2&17.9&5.4&61.0&66.3&47.3&33.1 \\ 

	    \multicolumn{1}{l}{CLIP \cite{radford2021learning} } &30.2&46.0&26.1&11.0&15.8&49.1&15.6&5.0 \\

		\multicolumn{1}{l}{COOP \cite{zhou2022learning}} &34.4&47.6&29.8&13.5&52.1&49.3&34.6&18.8\\
        
            \multicolumn{1}{l}{CoP \cite{zhang2024learning}} &47.0&\underline{50.9}&36.4&19.7&64.8&\underline{67.3}&\underline{51.2}&\underline{36.2}\\ 

            \multicolumn{1}{l}{ESE-GAN \cite{li2024ese}} &28.7&29.6&15.4&8.4&60.5&57.4&39.5&31.8\\
            
	    \multicolumn{1}{l}{Troika \cite{huang2024troika}} &\underline{48.8}&50.7&\underline{38.3}&\underline{20.8}&\underline{66.4}&61.2&47.8&33.0\\

	    \rowcolor{gray!15}\multicolumn{1}{l}{\textbf{SalientFusion}} &\textbf{50.5}{$\pm$\tiny .8}&\textbf{51.8}{$\pm$\tiny .8}&\textbf{39.4}{$\pm$\tiny .8}&\textbf{22.2}{$\pm$\tiny .8}&\textbf{66.9}{$\pm$\tiny .2}&\textbf{71.4}{$\pm$\tiny .5}&\textbf{51.6}{$\pm$\tiny .3}&\textbf{38.1}{$\pm$\tiny .1} \\ \hline
        
	\end{tabular}}

	\label{sota_general}
        \vspace{-5mm}
\end{table}

\textbf{Validation of General CZSL.} In order to further validate the generalization of our method, we conduct experiments on MIT-states and UT-Zappos, which are widely used dataset for general CZSL. In Tab.~\ref{sota_general}, \textit{SalientFusion} still achieves the SOTA results, demonstrating its effectiveness beyond the food. These results show that both the task of CZSFR and the proposed method are applicable to general CZSL tasks, confirming the method’s scalability and the task’s versatility.
\vspace{-4mm}

\section{Conclusion}

\label{conclusion}

In this paper, we explore the CZSFR and propose a novel framework named \text{SalientFusion}. The proposed SalientFormer and DebiasAT solve the problem of redundant information, role confusion and semantic bias. Our method achieves the best performance on two food datasets and a general dataset for the CZSL task. Our work represents a significant exploration of the CZSL task in the fine-grained domain, expanding the scope of research in the CZSL. Our work opens up a new approach for exploring the fine-grained domain for CZSL. In the future, we aim to construct food datasets with more diverse attributes and objects for the CZSFR task to promote real-world applications. Furthermore, we will design methods to address the complexity of objects in CZSFR.

\section{Acknowledgements}

This work was supported by grants from the Natural Science Foundation of China (72473148).

\bibliographystyle{splncs04}
\bibliography{mybibliography}
%





% LNCS Homepage, \url{http://www.springer.com/lncs}, last accessed 2023/10/25
% \end{thebibliography}

\end{document}